\documentclass[10pt,conference,switch]{IEEEtran}
\IEEEoverridecommandlockouts
\usepackage{amsmath,amssymb,amsfonts}
\usepackage{algorithmic}
\usepackage{graphicx}
\usepackage{textcomp}
\usepackage[table]{xcolor}
\def\BibTeX{{\rm B\kern-.05em{\sc i\kern-.025em b}\kern-.08em
    T\kern-.1667em\lower.7ex\hbox{E}\kern-.125emX}}

\usepackage{enumitem}

\usepackage{todonotes}
\usepackage[numbers, sort&compress]{natbib}
\usepackage{url}
\usepackage{booktabs}
\usepackage{multirow}
\usepackage{censor}
\usepackage{hyperref}
\usepackage{soul}

\usepackage{subcaption}
\usepackage{mathrsfs}
\newcommand{\email}[1]{\href{mailto:#1}{#1}}

\usepackage{tikz}
\usetikzlibrary{positioning}
\usetikzlibrary{shapes}
    
\tikzstyle{block} = [draw, fill=white, rectangle, minimum height=3em, minimum width=5em]
\tikzstyle{sum} = [draw, fill=white, circle, node distance=1cm]
\tikzstyle{pinstyle} = [pin edge={to-,thin,black}]
\tikzstyle{data}= [rectangle split,rectangle split parts=5,draw,text centered]

\begin{document}

\title{
Targeted Attack on GPT-Neo 
for the SATML Language Model Data Extraction Challenge
}

\author{
\IEEEauthorblockN{Ali Al-Kaswan}
\IEEEauthorblockA{\textit{Delft University of Technology} \\
Delft, The Netherlands \\
\email{a.al-kaswan@tudelft.nl}} 
\and
\IEEEauthorblockN{Maliheh Izadi}
\IEEEauthorblockA{\textit{Delft University of Technology} \\
Delft, The Netherlands \\
\email{m.izadi@tudelft.nl}}
\and
\IEEEauthorblockN{Arie van Deursen}
\IEEEauthorblockA{\textit{Delft University of Technology} \\
Delft, The Netherlands \\
\email{arie.vandeursen@tudelft.nl}}
}

\maketitle

\begin{abstract}
Previous work has shown that Large Language Models 
are susceptible to so-called data extraction attacks. 
This allows an attacker to extract a sample 
that was contained in the training data,
which has massive privacy implications.
The construction of data extraction attacks is challenging, 
current attacks are quite inefficient, 
and there exists a significant gap in the extraction capabilities of untargeted attacks and memorization. 
Thus, targeted attacks are proposed, which identify if a given sample from the training data, is extractable from a model.
In this work, we apply a targeted data extraction attack to the SATML2023 Language Model Training Data Extraction Challenge.\footnote{Language Models Training Data Extraction Challenge: \url{https://github.com/google-research/lm-extraction-benchmark}\label{lmext}}
We apply a two-step approach. 
In the first step, 
we maximise the recall of the model 
and are able to extract the suffix for 69\% of the samples. 
In the second step, 
we use a classifier-based Membership Inference Attack on the generations. 
Our AutoSklearn classifier achieves a precision of 0.841.
The full approach reaches a score of 0.405 recall at a 10\% false positive rate, 
which is an improvement of 34\% over the baseline of 0.301. 

\end{abstract}


\section{Introduction}
\label{introduction}
Language Models have recently become popular 
due to their ability to generate natural text 
and have been applied in various fields, 
such as Software Engineering~\cite{al2023extending,izadi2022codefill}.
However, neural language models trained on sensitive datasets
have been shown to memorize parts of their training data~\cite{carlini2021extracting, carlini2022membership, hu2022membershipSurvey}.
With a data extraction attack, an adversary can recover
individual training examples from the model's training
dataset. 

The ability to extract training data has massive privacy implications. Models which are trained using private datasets might be exposing their records. Models trained using publicly mined data might be violating the contextual integrity of the data of internet users~\cite{carlini2021extracting}. Recent work has found that developing robust attacks to extract training data is challenging. 

In this work, we focus on targeted attacks. We use the Language Model Training Data Extraction Challenge, to develop our attack. The benchmark provides a prefix of 50 tokens from the training data, we are tasked with predicting the next 50 tokens (suffix). The targeted model is GPT-Neo with $1.3$ billion parameters~\cite{gpt-neo}. 

We propose a two-stage attack strategy shown in~\autoref{fig:overview}. In the first stage, the model is prompted to predict multiple suffixes for the given prefix. In this step, we optimise for the recall of the model, i.e., for how many prefixes a correct suffix is generated. 

In the second step, we make use of a Membership Inference Attack to select the correct suffix from the set of candidates. In this case, precision is more important.
We find that contrastive search is the best decoding strategy to generate as many correct candidate suffixes as possible.
%
We design a successful attack based on a binary classifier (based on AutoSklearn) 
to classify the candidate suffixes.
With our attack, we are able to achieve a recall of 0.405 at a 10\% false positive rate, which is a 34\% improvement over the baseline of $0.301$.

\section{Membership Inference Attack Security Game}
\label{background}
In this section, we define a black-box membership inference attack using a security game inspired by \citeauthor{carlini2022membership}~\cite{carlini2022membership}. 

Given a challenger \(C\) and an adversary \(\mathscr{A}\), the game is defined as follows:
\begin{enumerate}
    \item The challenger samples a dataset \(D \subset \mathbb{D}\) and trains a model \(M_\theta \leftarrow Training_M(D)\) on the sampled dataset
    \item \(C\) samples a bit \(b \leftarrow \{0,1\}\). If \(b = 0\), \(C\) selects a training point \(x \in D\), otherwise \(C\) selects a training point \(x \in ((D \cup \mathbb{D}) - (D \cap \mathbb{D}))\). The point is then provided to \(\mathscr{A}\).
    \item \(\mathscr{A}\) is allowed query access to the model \(M_\theta\) and may perform any other polynomial time operations.
    \item \(\mathscr{A}\) outputs his prediction bit \(\hat{b} \leftarrow \{0,1\}\)
    \item If \(\hat{b} = b\), \(\mathscr{A}\) wins, otherwise \(C\) wins.
\end{enumerate}
In other words, the challenger randomly samples a subset \(D\) from the dataset \(\mathbb{D}\) and trains a model \(M_\theta\) on the subset. 
The adversary is then tasked with distinguishing samples 
that are and are not contained in the training data subset.
Note that the adversary does not have access to the underlying distribution of the data, and neither does the adversary have access to the base model \(M\), which makes training shadow models impossible. These limitations on the adversary also loosen the constraints on the model \(M\), which can be trained from scratch in step (1). Other attacks \cite{mireshghallah2022quantifying} require a functional base model \(M\) which is further fine-tuned on \(D\).





\vspace{10mm}
\section{Experimental Setup}

\subsection{Overview}
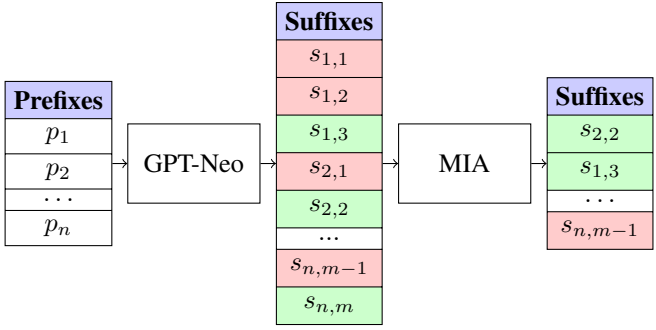
\begin{figure}[tb]
    \centering
    \begin{tikzpicture}[auto, node distance=1.80cm]
        \node [rectangle split, rectangle split parts=5,draw,text centered, , rectangle split part fill={blue!20, white}]
         (prefixes)
        {
            \textbf{Prefixes}
            \nodepart{two}
            \(p_1\)
            \nodepart{three}
            \(p_2\)
            \nodepart{four}
            \(\dots\)
            \nodepart{five}
            \(p_n\)
        };
        \node [block, name=generator, right of=prefixes] {GPT-Neo};
        \node [rectangle split, rectangle split parts=9,draw,text centered, right of=generator, rectangle split part fill={blue!20,red!20,red!20,green!20,red!20,green!20,white!20,red!20,green!20}]
         (suffixes)
        {
            \textbf{Suffixes}
            \nodepart{two}
            \(s_{1,1}\)
            \nodepart{three}
            \(s_{1,2}\)
            \nodepart{four}
            \(s_{1,3}\)
            \nodepart{five}
            \(s_{2,1}\)
            \nodepart{six}
            \(s_{2,2}\)
            \nodepart{seven}
            \(...\)
            \nodepart{eight}
            \(s_{n,m-1}\)
            \nodepart{nine}
            \(s_{n,m}\)
        };
        \node [block, name=mia, right of=suffixes] {MIA};
        \node [rectangle split, rectangle split parts=5,draw,text centered, right of=mia, rectangle split part fill={blue!20,green!20,green!20,white,red!20}]
         (results)
        {
            \textbf{Suffixes}
            \nodepart{two}
            \(s_{2,2}\)
            \nodepart{three}
            \(s_{1,3}\)
            \nodepart{four}
            \(\dots\)
            \nodepart{five}
            \(s_{n,m-1}\)
        };
        \draw [->] (prefixes) edge (generator) (generator) edge (suffixes) (suffixes) edge (mia) (mia) edge (results);
    \end{tikzpicture}
    \caption{An overview of the complete attack; From left to right, the prefixes are used by the GPT-Neo model to generate multiple suffixes per prefix, a MIA is then applied to select the presumed correct suffixes ordered by confidence.}
    \label{fig:overview}
\end{figure}

We show an overview of our attack in~\autoref{fig:overview}. 

\paragraph{Generation step}
We use the GPT-Neo model to generate suffixes for a given prefix. In the first step of the attack, we aim to increase the recall of the attack. We can generate multiple predictions per prefix, which will be filtered in the next step. 

It might be enticing to simply increase the number of predictions per prefix to get a higher chance of finding the right suffix. Doing this would increase the attack time and, more importantly, the number of errors in the MIA step. In the relative error-sensitive evaluation setting, this would be inadvisable.
\paragraph{MIA step}
In this step, we must infer which generated suffixes are members of the training data. In this step, we optimise for precision. For the sake of simplicity, we only select one sample per prefix. We also order the samples in descending order of confidence, such that the samples which are most probable to be correct are pushed up to the top. 
The metric we used to measure the performance of this step, and the total attack is the recall at a 10\% false positive rate. Concretely, this means that we count the number of correct predictions in the ordered output and stop counting when we count 10\% errors.

\subsection{Dataset}
The provided dataset consists of 15K samples. Each sample consists of a prefix and a suffix, both are 50 tokens long. The prefix prepended to the suffix is a 100-token sample from the Pile~\cite{pile}, an 800GB text dataset used to train GPT-Neo. The authors of the benchmark selected the samples such that for a given prefix, there is only a single unique suffix contained in the Pile~\cite{pile}.

As suggested by the authors of the benchmark, we use the first 14K samples to train and we isolate the last 1K samples for internal testing. Once we have obtained our solution we can test it with an additional 1K-sample validation set.

\section{Results}

\subsection{Generation Strategies}
Table \ref{fig:gen} shows the results for the different generation strategies. We ran the GPT-Neo model with different decoding settings on 100 prefixes. We prompt the model to generate several different generations per prefix.

We used the Greedy, Contrastive, and Beam decoding strategies. We first ran the different generation strategies to generate 10 generations per prefix, on the standard settings. We found that contrastive search obtains the highest recall of the tested stratagems.

Further testing with different settings for penalty\_alpha and top\_k, 
shows that the standard settings have the highest recall for ten generations.

Furthermore, we found that the recall of beam search decreased once we increased the beam size above 10 beams. Overall, we found beam search with a sufficiently large beam size to compete with Contrastive search to be too slow and memory intensive to use.

\begin{table}[!tb]
    \centering
    \begin{tabular}{lll|l}
    \noalign{\smallskip}\toprule
    Strategy & Settings & Generations& Recall  \\ 
    \cmidrule{1-4}
    Greedy          & p=1,k=10  &  10       &  \textbf{0.50}       \\ 
    \cmidrule{1-1}
    Contrastive     & a=0.6,k=4 &  1        &  0.28       \\ 
                    & a=0.6,k=4 &  10       &  \textbf{0.58}       \\ 
                    & a=0.6,k=2 &  10       &  0.52       \\ 
                    & a=0.9,k=4 &  10       &  0.49       \\ 
                    & a=0.2,k=4 &  10       &  0.48       \\ 
                    & a=0.9,k=10 &  10       &  0.46       \\ 
                    & a=0.6,k=4 &  100      &  \textbf{0.69}       \\
    \cmidrule{1-1}
    Beam            & beam=50   &  3        &  0.57       \\ 
                    & beam=10   &  10       &  \textbf{0.67}       \\ 
                    & beam=25   &  25       &  0.53       \\ 
    \noalign{\smallskip}\bottomrule
    \end{tabular}
    \caption{Recall per decoding strategy for 100 prefixes}
    \label{fig:gen}
\end{table}

\begin{figure}[tb]
    \centering
    \includegraphics[width=\linewidth]{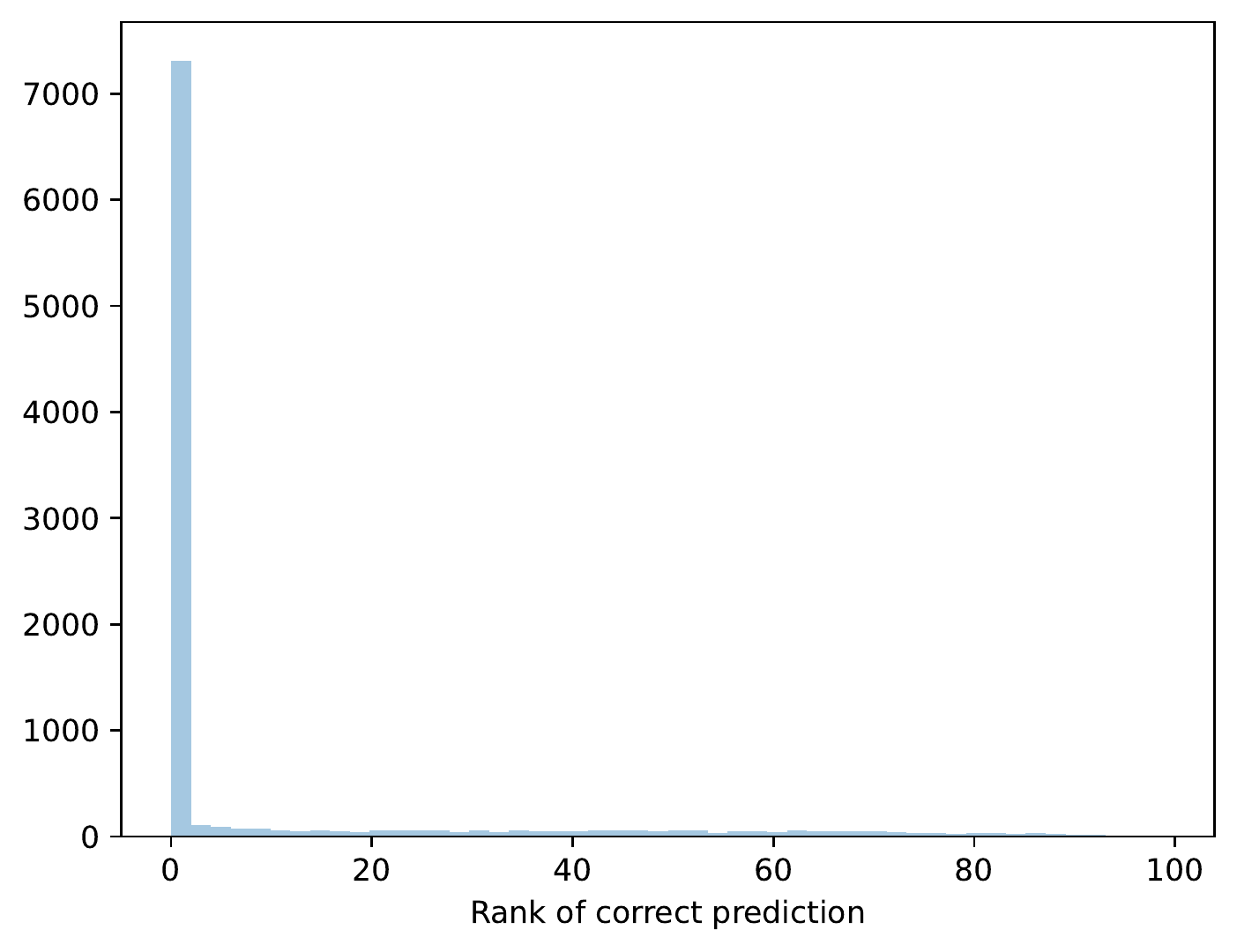}
    \caption{Rank of correct prediction (if exists)}
    \label{fig:rank}
\end{figure}

Finally, we use GPT-Neo with the best generation strategy, namely, contrastive search \(\alpha = 0.6, k = 4\) with $100$ generations per prefix and plot the rank of the correct prediction in Figure~\ref{fig:rank}. The generations are ranked by the model loss on the generation. Note, that we omit the prefixes for which the model was unable to generate the correct prefix. This figure shows that if the correct prefix is available, it is usually the one with the lowest loss. The remaining challenge is to distinguish between the prefixes which have and the prefixes which do not have a correct suffix associated with them. 

\subsection{Classification MIA}
We train several classifiers on the task of distinguishing between members and non-members. We first use GPT-Neo with the best generation strategy, namely, contrastive search \(\alpha = 0.6, k = 4\) with 100 generations per prefix. We apply this to the entire dataset of 15K prefixes. We apply a filter and only consider the samples with the lowest loss for each prefix, we found that this improves the attack, and reduces the computational costs. We split the data and use the first 14K as training data and the last 1K as a test set. The recall of the generation step on the test set was 0.669, which is in line with our previous findings. After filtering this was reduced to 0.498.

We use the Sklearn~\cite{scikit-learn} implementation of the standard classifiers. The classifiers were trained until convergence. The AutoSklearn~\cite{feurer2020autosklearn} classifier was trained for 10 minutes, 60 seconds per model, with 16 threads. For tokenization, we use the standard Sklearn TF-IDF pipeline and the Sentence-Transformers package with the 'all-mpnet-base-v2' model. We chose this model because it is the highest-performing one in the sentence embedding benchmark.~\footnote{Sentence Embedding Benchmark: \url{https://www.sbert.net/docs/pretrained\_models.html}}

\begin{figure}[tb]
    \centering
    \includegraphics[width=\linewidth]{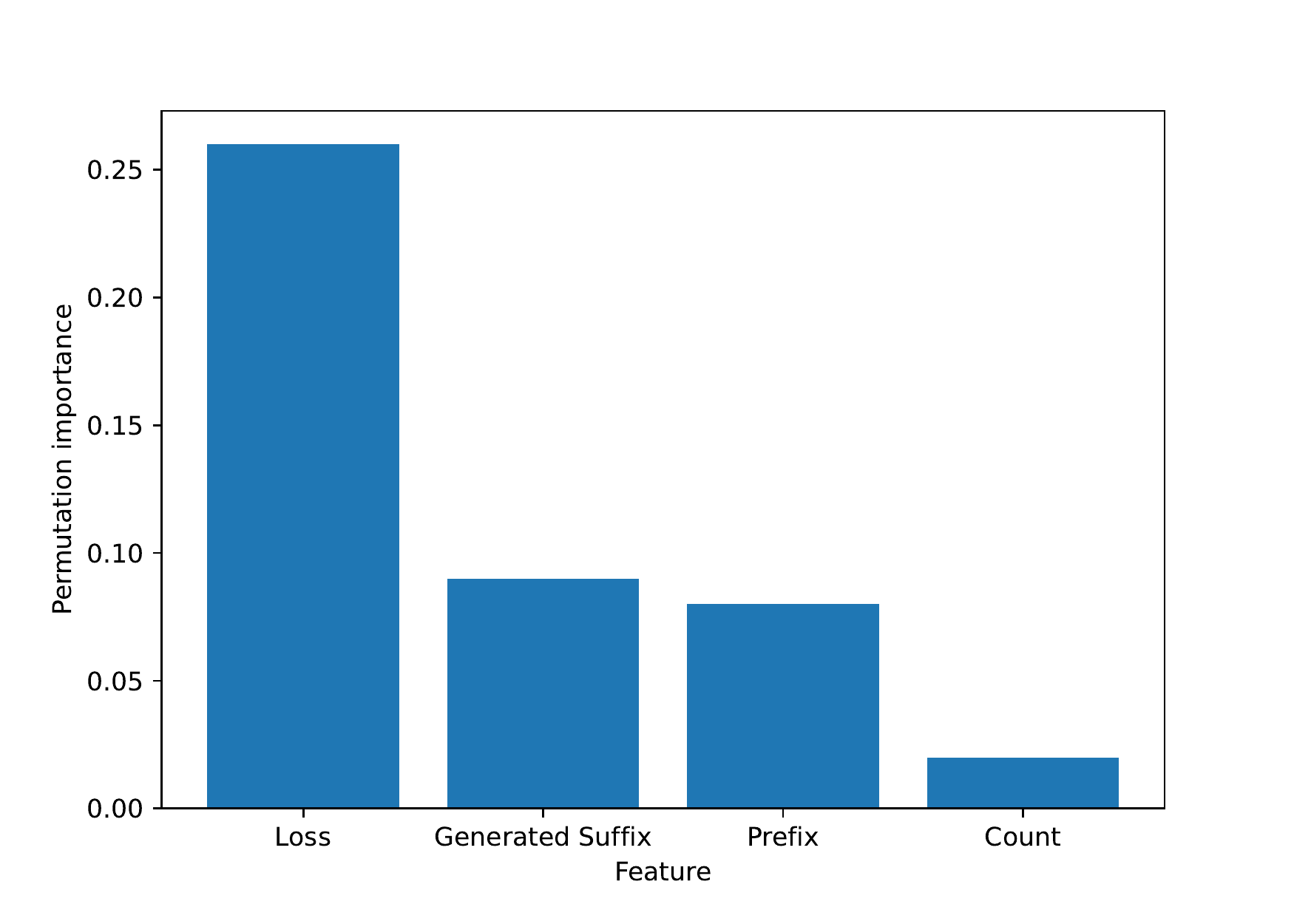}
    \caption{Permutation importance of features}
    \label{fig:feature}
\end{figure}

Besides the prefix and generated suffix, we also include the number of unique generations produced by the model (count), as well as the model loss as features. We plot the permutation importance of the different features to our AutoSklearn model performance in~\autoref{fig:feature}.
We found that the loss is by far the most important feature, while the textual features do contribute to the performance, their importance is limited. Finally, 
the number of distinct generations has a minimal contribution to the performance.

We tested Logistic Regression, Stochastic Gradient Descent with both Huber and perceptron losses, Support Vector Machines, Gaussian Naive Bayes, and Gradient Boost models. 

To get the final output of the attack, we simply sort the samples by the probability estimate of the model. For the models that cannot calculate a probability estimate, we apply a filter to remove the predicted non-members and we order the samples by the loss.

To score the solutions,
we opted to use precision as this attack values a low false positive rate. 
Furthermore, we also calculate the final accuracy of the attack through the precision at a 10\% false positive rate. Note that, the maximum achievable score is limited by the recall of the previous step, namely a recall of 498 at a 10\% false positive rate. 

\begin{table}[tb]
    \centering
    \begin{tabular}{ll|ll}
    \noalign{\smallskip}\toprule
    Feature Extraction   & Strategy      &   Precision   &   R@10\%FPR \\ \cmidrule{1-4}
    Baseline        &     -         &   -       &   0.301\\
    \cmidrule{1-4}
    AutoSklean      & - & \textbf{0.841}& \textbf{0.405}\\ 
    \cmidrule{1-1}
    TF-IDF          & Log Reg       &   0.808   &   0.397 \\ 
                    & SGD huber     &   0.520   &   0.097  \\
                    & SGD perceptron&   0.784   &   0.302 \\
                    & SVM           &   0.639   &   0.279 \\
                    & GaussianNB    &   0.599   &   0.273 \\
                    & Gradient Boost&   0.766   &   0.365 \\
    \cmidrule{1-1}
    S-Transformers  & Log Reg       &   0.780   &   0.345 \\
                    & SGD huber     &   0.466   &   0.279 \\
                    & SGD perceptron&   0.602   &   0.280 \\
                    & SVM           &   0.498   &   0.126 \\
                    & GaussianNB    &   0.608   &   0.231 \\
                    & Gradient Boost&   0.776   &   0.359 \\
                    & AutoSklearn   &   0.807   &   0.397 \\ 

    \noalign{\smallskip}\bottomrule
    \end{tabular}
    \caption{Precision and overall attack score on test set}
    \label{fig:classification}
\end{table}

AutoSklearn, with its automatic feature extraction pipeline, performs best. We found that further increasing the training time, does not improve the models' performance. We found that halving the training time, gave a slight decrease in performance. The actual convergence point lies somewhere between 5 and 10 minutes. Furthermore, all of the models in the constructed ensemble are Gradient Boost models.

The second best is a tie between logistic Regression with TF-IDF and AutoSklearn with Sentence-Transformers. Note that while AutoSklearn takes around 10 minutes to train, Logistic Regression only takes around three seconds.

We further found that the Sentence-Transformers embeddings are of a much lower dimensionality than TF-IDF (768 vs 28182). We were therefore unable to load the TF-IDF embeddings into AutoSklearn. This reduction greatly speeds up the training process, but the models perform slightly worse. This small difference can be explained by the fact that Sentence-Transformers are Deep-Learning based embedding models, which take the semantic meaning of a sentence in mind, while TF-IDF is a simple statistical method. 

\subsection{Validation Scores}
\begin{figure}[tb]
    \centering
    \includegraphics[width=\linewidth]{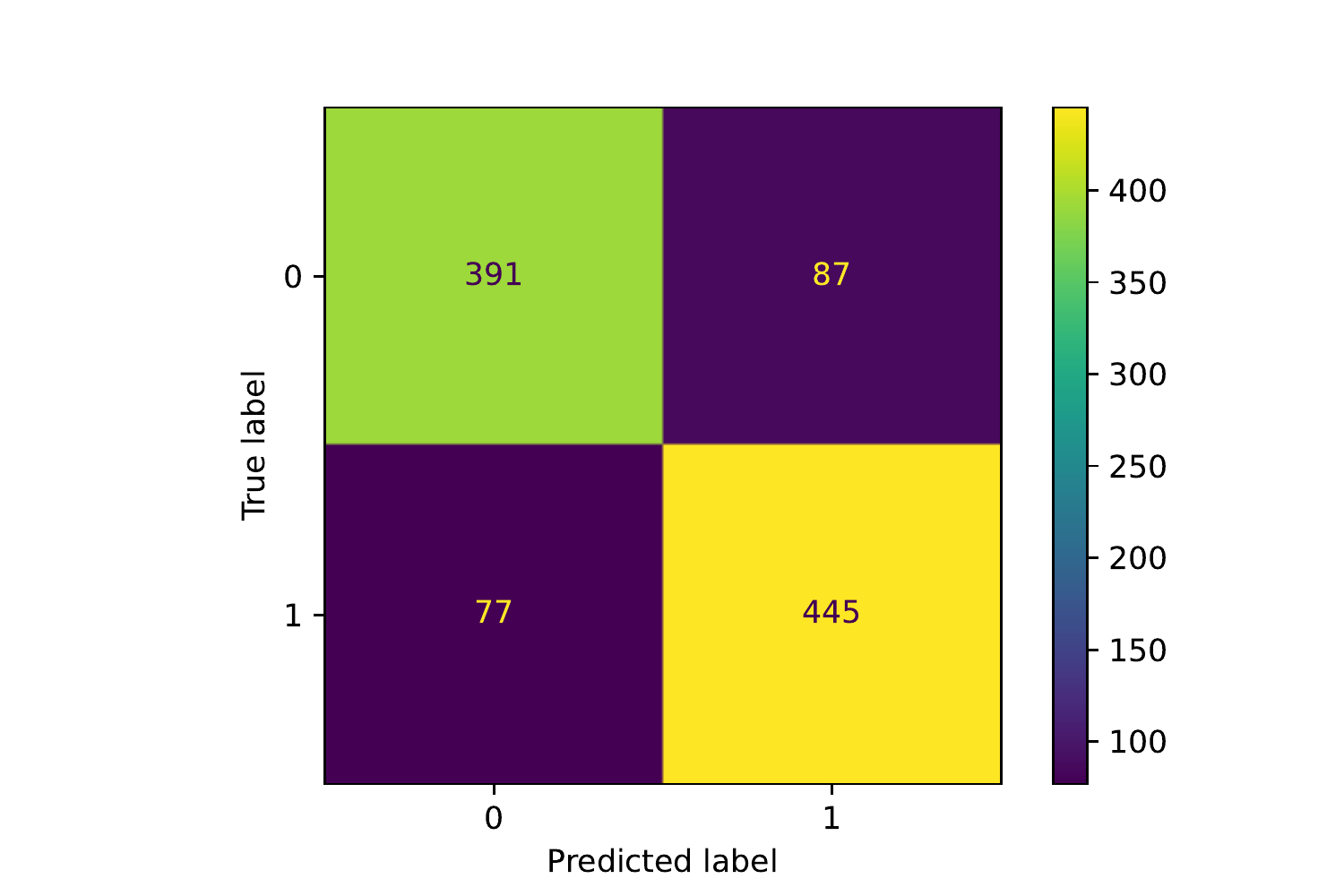}
    \caption{Confusion matrix of validation set}
    \label{fig:confusion}
\end{figure}

We finally run the trained AutoSklearn model on the validation set provided by the organizers. The final score on the validation set is a recall of \textbf{0.413} at a 10\% false positive rate. \autoref{fig:confusion} shows the confusion matrix on the validation set, which shows that the model is quite balanced in its predictions, and does not heavily favour precision or recall while achieving high accuracy. 

\section{Discussion}

The proposed attack is relatively quick to run, as it does not require any type of fine-tuning or prompt-tuning. The slowest aspect of the attack is the generation step, to generate 100 candidate samples for 1K prefixes, we require around 1 hour on an Nvidia RTX 3080, the MIA itself runs in a few seconds. This is around the same speed as the baseline attack.

With our classification-based membership inference attack, we seem to have relatively high precision. We believe that we are approaching the limit set by the generation step. Recall that after generating and filtering, we only extracted the correct suffix for 49.8\% of the samples. This indicates that there is still much room for improvement in the generation step of our proposed attack. 
We only investigated different decoding strategies and did not alter the prefixes. Prompt engineering or prefix-tuning might increase the recall of the generation step and therefore the score of the entire attack.

Another possible improvement is to introduce more features into the classifier. Instead of using a different method to create an embedding for the textual features, we can use the embedding vector produced by the GPT-Neo model, this would however turn the attack into a white-box variant. 

\section{Conclusion}
To conclude, we proposed a novel two-phased attack strategy. 
In the first step, we find the best decoding strategy 
to maximise the recall of the attack. 
In the second step, we use a binary classifier to select the best suffix. 
Our approach was able to show an improvement of 34\% 
over the baseline score with minimal additional runtime requirements over the provided baseline.

\bibliographystyle{IEEEtranN}
\bibliography{references.bib}

\begin{thebibliography}{10}
\providecommand{\natexlab}[1]{#1}
\providecommand{\url}[1]{#1}
\csname url@samestyle\endcsname
\providecommand{\newblock}{\relax}
\providecommand{\bibinfo}[2]{#2}
\providecommand{\BIBentrySTDinterwordspacing}{\spaceskip=0pt\relax}
\providecommand{\BIBentryALTinterwordstretchfactor}{4}
\providecommand{\BIBentryALTinterwordspacing}{\spaceskip=\fontdimen2\font plus
\BIBentryALTinterwordstretchfactor\fontdimen3\font minus
  \fontdimen4\font\relax}
\providecommand{\BIBforeignlanguage}[2]{{%
\expandafter\ifx\csname l@#1\endcsname\relax
\typeout{** WARNING: IEEEtranN.bst: No hyphenation pattern has been}%
\typeout{** loaded for the language `#1'. Using the pattern for}%
\typeout{** the default language instead.}%
\else
\language=\csname l@#1\endcsname
\fi
#2}}
\providecommand{\BIBdecl}{\relax}
\BIBdecl

\bibitem[Al-Kaswan et~al.(2023)Al-Kaswan, Ahmed, Izadi, Sawant, Devanbu, and
  van Deursen]{al2023extending}
A.~Al-Kaswan, T.~Ahmed, M.~Izadi, A.~A. Sawant, P.~Devanbu, and A.~van Deursen,
  ``Extending source code pre-trained language models to summarise decompiled
  binaries,'' in \emph{Proceedings of the 30th IEEE International Conference on
  Software Analysis, Evolution and Reengineering (SANER)}, 2023.

\bibitem[Izadi et~al.(2022)Izadi, Gismondi, and Gousios]{izadi2022codefill}
M.~Izadi, R.~Gismondi, and G.~Gousios, ``Codefill: Multi-token code completion
  by jointly learning from structure and naming sequences,'' in
  \emph{Proceedings of the 44th International Conference on Software
  Engineering (ICSE)}.\hskip 1em plus 0.5em minus 0.4em\relax ACM, 2022, p.
  401–412.

\bibitem[Carlini et~al.(2021)Carlini, Tramer, Wallace, Jagielski, Herbert-Voss,
  Lee, Roberts, Brown, Song, Erlingsson, et~al.]{carlini2021extracting}
N.~Carlini, F.~Tramer, E.~Wallace, M.~Jagielski, A.~Herbert-Voss, K.~Lee,
  A.~Roberts, T.~Brown, D.~Song, U.~Erlingsson \emph{et~al.}, ``Extracting
  training data from large language models,'' in \emph{30th USENIX Security
  Symposium (USENIX Security 21)}, 2021, pp. 2633--2650.

\bibitem[Carlini et~al.(2022)Carlini, Chien, Nasr, Song, Terzis, and
  Tramer]{carlini2022membership}
N.~Carlini, S.~Chien, M.~Nasr, S.~Song, A.~Terzis, and F.~Tramer, ``Membership
  inference attacks from first principles,'' in \emph{2022 IEEE Symposium on
  Security and Privacy (SP)}.\hskip 1em plus 0.5em minus 0.4em\relax IEEE,
  2022, pp. 1897--1914.

\bibitem[Hu et~al.(2022)Hu, Salcic, Sun, Dobbie, Yu, and
  Zhang]{hu2022membershipSurvey}
H.~Hu, Z.~Salcic, L.~Sun, G.~Dobbie, P.~S. Yu, and X.~Zhang, ``Membership
  inference attacks on machine learning: A survey,'' \emph{ACM Computing
  Surveys (CSUR)}, vol.~54, no. 11s, pp. 1--37, 2022.

\bibitem[Black et~al.(2021)Black, Gao, Wang, Leahy, and Biderman]{gpt-neo}
\BIBentryALTinterwordspacing
S.~Black, L.~Gao, P.~Wang, C.~Leahy, and S.~Biderman, ``{GPT-Neo: Large Scale
  Autoregressive Language Modeling with Mesh-Tensorflow},'' Mar. 2021, {If you
  use this software, please cite it using these metadata.} [Online]. Available:
  \url{https://doi.org/10.5281/zenodo.5297715}
\BIBentrySTDinterwordspacing

\bibitem[Mireshghallah et~al.(2022)Mireshghallah, Goyal, Uniyal,
  Berg-Kirkpatrick, and Shokri]{mireshghallah2022quantifying}
F.~Mireshghallah, K.~Goyal, A.~Uniyal, T.~Berg-Kirkpatrick, and R.~Shokri,
  ``Quantifying privacy risks of masked language models using membership
  inference attacks,'' \emph{arXiv preprint arXiv:2203.03929}, 2022.

\bibitem[Gao et~al.(2020)Gao, Biderman, Black, Golding, Hoppe, Foster, Phang,
  He, Thite, Nabeshima, Presser, and Leahy]{pile}
L.~Gao, S.~Biderman, S.~Black, L.~Golding, T.~Hoppe, C.~Foster, J.~Phang,
  H.~He, A.~Thite, N.~Nabeshima, S.~Presser, and C.~Leahy, ``The {P}ile: An
  800gb dataset of diverse text for language modeling,'' \emph{arXiv preprint
  arXiv:2101.00027}, 2020.

\bibitem[Pedregosa et~al.(2011)Pedregosa, Varoquaux, Gramfort, Michel, Thirion,
  Grisel, Blondel, Prettenhofer, Weiss, Dubourg, Vanderplas, Passos,
  Cournapeau, Brucher, Perrot, and Duchesnay]{scikit-learn}
F.~Pedregosa, G.~Varoquaux, A.~Gramfort, V.~Michel, B.~Thirion, O.~Grisel,
  M.~Blondel, P.~Prettenhofer, R.~Weiss, V.~Dubourg, J.~Vanderplas, A.~Passos,
  D.~Cournapeau, M.~Brucher, M.~Perrot, and E.~Duchesnay, ``Scikit-learn:
  Machine learning in {P}ython,'' \emph{Journal of Machine Learning Research},
  vol.~12, pp. 2825--2830, 2011.

\bibitem[Feurer et~al.(2020)Feurer, Eggensperger, Falkner, Lindauer, and
  Hutter]{feurer2020autosklearn}
M.~Feurer, K.~Eggensperger, S.~Falkner, M.~Lindauer, and F.~Hutter,
  ``Auto-sklearn 2.0: The next generation,'' \emph{arXiv preprint
  arXiv:2007.04074}, vol.~24, 2020.

\end{thebibliography}
\end{document}